\documentclass[conference]{IEEEtran}
\IEEEoverridecommandlockouts
\usepackage{cite}
\usepackage{amsmath,amssymb,amsfonts}
\usepackage{algorithmic}
\usepackage{graphicx,booktabs,multirow}
\usepackage{textcomp}
\usepackage{xcolor, soul}
\usepackage{enumitem}
\usepackage{url}
\usepackage{fancyvrb}
\usepackage{commath}
\usepackage{comment}
\usepackage{fleqn,multirow,wrapfig,lineno,hyperref}
\hypersetup{
    colorlinks=true,
    linkcolor=blue,
    filecolor=blue,      
    urlcolor=blue,
    bookmarksopen=true,
    citebordercolor=blue,
    filebordercolor=blue,
    linkbordercolor=blue,
    menubordercolor=blue,
    urlbordercolor=blue
}
\sethlcolor{yellow}

\definecolor{bronze}{rgb}{0.8, 0.5, 0.2}
\definecolor{blue}{rgb}{0, 0, 1}
\definecolor{green}{rgb}{0, 1, 0}
\definecolor{black}{rgb}{0, 0, 0}
\definecolor{red}{rgb}{1, 0, 0}

\newcommand{\myVspace}{\vspace{5 pt}}

\graphicspath{{figs/}}

\def\BibTeX{{\rm B\kern-.05em{\sc i\kern-.025em b}\kern-.08em
    T\kern-.1667em\lower.7ex\hbox{E}\kern-.125emX}}

\begin{document}

% \title{
% Efficient and Safe Physical Human-Robot Interaction Using Adaptive Admittance Control via Learning-based Subtask Detection and\\Progress Estimation
% }

\title{
\vspace{10pt}
%Efficient and Safe Physical Human-Robot Interaction 
%for Contact-Rich Manufacturing Tasks
%via Learning-based Motion Estimation
Estimating Human Muscular Fatigue \\
in Dynamic Collaborative Robotic Tasks \\
with Learning-Based Models
}

% Authors and affiliations
\author{Feras~Kiki$^{1}$, Pouya~P.~Niaz$^{1}$,
        Alireza~Madani$^{1}$,
        and Cagatay~Basdogan$^{1}$% <-this % stops a space
\thanks{$^{1}$ F. Kiki, P. P. Niaz, A. Madani, and C. Basdogan (\textit{corresponding author}) are with the Robotics and Mechatronics Laboratory (RML) and the KUIS AI Center, Koc University, Sariyer, Istanbul 34450, Turkey.
{\tt\small {\{fkiki18, pniaz20, amadani20, cbasdogan\}@ku.edu.tr}}% <-this % stops a space
}
}

\maketitle

\begin{abstract}
Assessing human muscle fatigue is critical for optimizing performance in physical human–robot interaction (pHRI) tasks and mitigating safety risks for the human operator. This paper presents a data-driven framework for estimating muscle fatigue in dynamic pHRI tasks using surface electromyography (sEMG) sensors attached to the human arm. Subject-specific machine learning (ML) regression models were developed to estimate fatigue levels during cyclic (i.e., repetitive) pHRI tasks. Specifically, Random Forest, XGBoost, and Linear Regression models were trained to estimate the fraction of cycles to fatigue (FCF) using three frequency-domain and one time-domain EMG features. Their performance was benchmarked against a convolutional neural network (CNN) that processes spectrogram representations of filtered EMG signals. Unlike most earlier data-driven approaches that primarily formulated fatigue estimation as a classification problem, our method models the progression toward fatigue through regression, enabling tracking of gradual physiological changes rather than discrete states, which is critical for timely intervention and adaptive control in dynamic pHRI tasks.
Experiments were conducted with ten participants who interacted with a collaborative robot driven by an admittance controller, performing lateral (left-right) cyclic movements of the end effector until the onset of muscular fatigue. The results demonstrate that the root mean square error (RMSE) of FCF estimation across participants was 20.8 ± 4.3\%, 23.3 ± 3.8\%, 24.8 ± 4.5\%, and 26.9 ± 6.1\% for the CNN, Random Forest, XGBoost, and Linear Regression models, respectively. To examine cross-task generalization in this investigational study, additional experiments were performed with one participant who executed vertical (up–down) and circular repetitive movements. Models trained solely on the lateral-movement data were directly tested on these unseen tasks. The results indicate that the proposed ML/DL models are robust to variations in movement direction, arm kinematics, and muscle recruitment patterns, while the Linear Regression model performed poorly. 
\end{abstract}

\begin{IEEEkeywords}
Physical human-robot interaction, muscle fatigue estimation, machine learning, ergonomics and human factors, collaborative robotics, surface electromyography (sEMG), and admittance control.
\end{IEEEkeywords}

\section{Introduction}
    \label{sec:introduction}
The field of physical human-robot interaction (pHRI) has grown significantly during the last two decades. Since the human is an integral part of pHRI, improving human comfort and ergonomics is important not only for maximizing productivity and efficiency in collaborative tasks \cite{guler2022adaptive, madani2022robot} but also for reducing safety risks to the human operator \cite{niaz2024learning}, including the potential adverse effects of muscle fatigue \cite{Peternel2016, peternel2018robot}, which can impair performance and increase injury risk over time. 
Fatigue is a physiological state that takes place after prolonged/accumulated high effort. If the progression of human effort is estimated during a pHRI task, the robot's contribution can be dynamically adjusted to reduce the load on the human and prevent the occurrence of muscle fatigue \cite{Zahedi2021}. For example, in the manufacturing domain, estimating the progression of human effort helps to evaluate the difficulty level of the collaborative task and its sub-tasks to make changes toward reducing potential musculoskeletal injuries. 
Fatigue estimation is also essential in rehabilitation and sports training to prevent overexertion, tailor exercise intensity, and ensure optimal performance and safe recovery. For example, robotic rehabilitation for post-stroke patients often involves repetitive, rhythmic movements designed to help restore damaged neural pathways \cite{mashayekhi2022emg}. However, the onset of fatigue can interrupt these training sessions and restrict the overall duration of therapy. Therefore, finding a systematic and accurate approach to estimating the progression of human effort towards fatigue is considered to be a knowledge gap in the state-of-the-art.

\begin{figure}[t]
    \myVspace
	\includegraphics[width=\columnwidth]{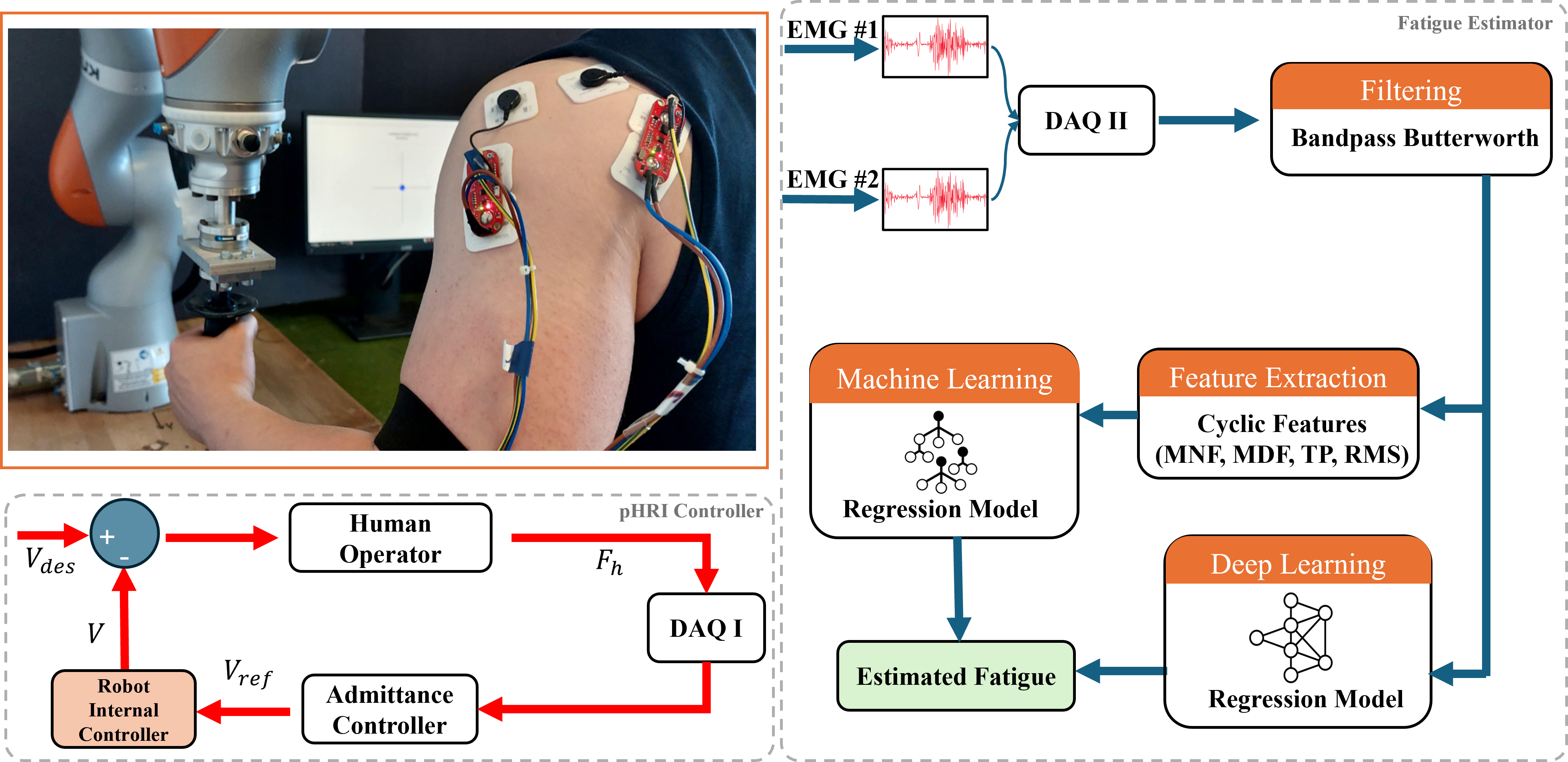}
	\centering
	\caption{Our approach to estimating human muscular fatigue in cyclic pHRI tasks employs learning-based regression models that predict the fraction of cycles to fatigue (FCF) from sEMG measurements.} 
	\label{fig:our_approach}
\end{figure}
In some of the earlier pHRI studies, human muscular effort has been indirectly inferred from the force applied by the user, which is typically acquired by a force sensor. Nonetheless, measuring the force alone often does not provide a complete assessment of human muscular effort in various scenarios. For instance, in collaborative drilling \cite{peternel2019selective, sirintuna2020variable, madani2022robot} and collaborative surface polishing \cite{peternel2018robot, hamdan2024robotic}, the magnitude of the normal force applied by the human operator to the workpiece may remain constant, yet muscle fatigue could still occur due to prolonged effort. In both cases, the muscle activation can vary over time due to the dynamic nature of the task.  
Hence, not just the force applied by the human, but muscular effort should also be assessed by monitoring muscle activity, which is typically achieved by using electromyography (EMG) sensors \cite{peternel2018robot, peternel2019selective}. EMG sensors detect the electrical signals generated by muscle contractions and can provide information about the intensity and timing of muscle activation. Prolonged efforts in pHRI may lead to a progressive decline in a muscle's ability to sustain performance as observed by the changes in EMG signals, resulting in muscle fatigue, which should be avoided to reduce musculoskeletal injuries and potential disorders.

\section{Related Literature}
\label{sec:Related Literature}
De Luca\cite{de1997use} identified frequency-based features such as mean power frequency (MNF) and median power frequency (MDF) as reliable metrics for assessing muscle fatigue based on surface EMG measurements. As muscle fatigue occurs, the power spectrum shifts to lower frequencies, and hence, the MNF and MDF decrease gradually. It has also been reported that the RMS amplitude of the EMG signal increases towards the onset of fatigue.

Although these features are effective indicators of fatigue in static loading tasks (as in holding a heavy object in the mid-air) involving isometric muscle contractions, they are less pronounced in dynamic loading tasks (as in repeatedly moving a heavy object up and down), involving isotonic contractions \cite{gonzalez2012electromyographic, chand2023dynamic}. In a dynamic task, muscles cyclically shorten (concentric phase) and lengthen (eccentric phase) while generating force, promoting intermittent relaxation and improved blood circulation, which helps delay fatigue compared to static loading \cite{mugnosso2019coupling, shariatzadeh2023predicting, chand2023dynamic}. Fernando et al. \cite{fernando2016estimation} observed that the ratio of MNF to the average rectified value (ARV) provides a more accurate measure of fatigue in dynamic tasks. However, their results were not consistent across subjects. Zahedi et al. \cite{Zahedi2021} relied on the integral of the normalized EMG signal to estimate overall human effort in a dynamic reaching task performed with a robot. They adjusted the robot's damping based on the effort required, ensuring that the subjects did not experience muscle fatigue. Peternel et al. \cite{Peternel2016} incorporated EMG measurements into a first-degree ordinary differential equation (ODE) to estimate muscle fatigue in a cyclic sawing task performed with a robot. However, this approach does not account for muscle recovery, which was later investigated by Lorenzini et al. \cite{lorenzini2019new}. To incorporate the effect of dynamic muscle contractions, including both concentric (shortening) and eccentric (lengthening) actions, of the human arm into the effort estimation, Peternel et al. \cite{peternel2018online} and Figueredo et al. \cite{figueredo2021planning} employed a musculoskeletal model developed in OpenSim \cite{seth2018opensim}, an open-source software platform for modeling, simulating, and analyzing human musculoskeletal system and movement. 

\begin{comment}
Machine learning (ML) models have been used in the past for classification of human action and intent recognition based on the features (see the review in \cite{simao2019review}). More recently, focus has shifted towards deep learning (DL) models since they do not require any feature engineering (see the review in \cite{xiong2021deep}). For example, Su et al. \cite{su2021deep} utilized a convolutional neural network (CNN) model to map the features extracted from EMG signals automatically to the interaction force between human and robot. 
\end{comment}

More recently, the research in this domain has focused on data-driven machine learning (ML) and deep learning (DL) models for fatigue estimation. Initial studies employing ML models based on EMG measurements primarily targeted the classification of human actions and intent recognition, rather than fatigue \cite{simao2019review}). Subsequent efforts have shifted toward DL models, which eliminate the need for manual feature engineering \cite{xiong2021deep}. In the context of fatigue estimation, for example, Karthick et al. \cite{karthick2018surface} evaluated five ML classifiers utilizing 12 EMG features to distinguish between dynamic muscle fatigue and non-fatigue conditions during biceps curl exercises with a 6 kg load. Moniri et al. \cite{moniri2020real} employed a deep CNN model to simultaneously learn and predict five common EMG features in real time for assessing fatigue in trunk muscles.

Fatigue estimation using EMG-driven ML/DL models has received relatively little attention within the domain of pHRI. In \cite{peternel2019selective}, due to the musculoskeletal model's requirement for intensive computations and its limitation to offline operation, a Gaussian Process Regression (GPR) machine learning model was implemented to create a real-time, configuration-dependent mapping from the endpoint force of the robot to the internal muscle forces to estimate human muscular fatigue. You et al. \cite{you2024human} developed a digital twin (DT) framework for real-time fatigue estimation in manufacturing. The DT utilizes a musculoskeletal model developed in OpenSim \cite{seth2018opensim}, a bidirectional Long Short-Term Memory (BiLSTM) network, and inverse kinematics to estimate muscle forces first, and then the fatigue based on the ODE model proposed in \cite{peternel2018online}. 

\begin{comment}
For example, Sirintuna et al. \cite{sirintuna2020detecting} utilized an artificial neural network (ANN) model to predict the intended direction of human movement by utilizing electromyography (EMG) signals in a collaborative drilling task. They employed this classifier in an admittance control architecture to constrain human arm motion to the intended direction and prevent undesired movements along other directions. In \cite{peternel2019selective}, due to the musculoskeletal model's requirement for intensive computations and its limitation to offline operation, a Gaussian Process Regression (GPR) machine learning model was implemented to create a real-time, configuration-dependent mapping from the endpoint force to the internal muscle forces to estimate human muscular fatigue. Su et al. \cite{su2021deep} developed a deep learning (DL) model to map EMG features to the interaction force between human and robot. A convolutional neural network (CNN) was used to extract features automatically from the EMG signals without using a musculoskeletal model.
\end{comment}

In this study, we introduce a learning-based approach to estimate the progression towards human muscular fatigue in pHRI for dynamic and repetitive tasks by analyzing electromyography (EMG) signals acquired from arm muscles (see Fig.~\ref{fig:our_approach}). The contributions of our study are as follows 
\begin{itemize}
\item We conducted physical human–robot interaction (pHRI) experiments in virtual environments (VEs) using a collaborative robotic arm operated under 3-degree-of-freedom (3-DoF) admittance control. Surface electromyography (EMG) signals were recorded from the arm muscles of ten human participants to train subject-specific ML and DL models for detecting muscular fatigue. The robot's control loop operated at a high update frequency, enabling real-time haptic rendering of interaction forces during task execution. This real-time force feedback capability within the VE framework facilitated the design and execution of complex pHRI tasks that would be challenging or impractical to implement in physical settings \cite{niaz2024learning}, thereby providing a flexible and safe platform for collecting high-quality data for training the learning models. 
\item Using the learning models, we estimate the fraction of cycles to fatigue (FCF) for cyclic and repetitive pHRI tasks. Previous data-driven ML/DL approaches to fatigue estimation predominantly framed the problem as classification, whereas our approach models the continuous progression of fatigue through regression. This formulation enables tracking of gradual physiological changes rather than discrete states, which is critical for timely intervention and adaptive control in dynamic tasks. We also validate our FCF estimates with the human operator's self-rating of perceived fatigue for each cycle during the task, which are found to be strongly correlated.
\item To enhance the generalizability of our approach across different movement kinematics and muscle groups, the ML regression models were trained using cycle-wise normalized EMG features with respect to the first cycle. The regression performances of the ML models were benchmarked against a DL model trained on spectrogram representations of the EMG signals. Following rigorous training and cross-validation for a single pHRI task, all trained models were evaluated on two additional, distinct pHRI tasks. This cross-task evaluation was designed to assess the robustness and generalization capability of the fatigue estimation framework under different interaction conditions.
\end{itemize}

\section{Approach}
    \label{sec:approach}
This section introduces our hardware setup for pHRI, our experiments, and the learning pipeline for estimating human muscular fatigue. 

\subsection{Hardware Setup}
    \label{sec:hardware}

Our pHRI system is composed of the following major hardware components: a collaborative robot (LBR iiwa 7 R800, KUKA Inc.), a force sensor (Mini45, ATI Inc.) for measuring the human force, and 2 EMG sensors (AT-04-001, MyoWare Inc.) attached to the user's arm for measuring the muscle activation during the execution of the pHRI task.

In our system, the data from the force sensor was sampled at 500 Hz by an internal DAQ card (PCI-6225, National Instruments Inc.) connected to a personal computer. The end-effector position and velocity of the robot are also acquired at the same update rate using the robot's encoders. The EMG data were acquired at 2 kHz by an external DAQ card (USB-6251, National Instruments Inc.) connected to the computer. All software code was developed in the C/C++ and Python programming languages, leveraging multi-threading techniques to enable the seamless acquisition, processing, and integration of multimodal data, including force, robot kinematics, and EMG.

\subsection{Closed Loop Control Architecture}
    \label{sec:controlsystem}

We employ three decoupled admittance controllers, each dedicated to one of the three Cartesian degrees of freedom (dof), to manage the interactions between the human operator and robot. Our control architecture is shown in Fig.~\ref{fig:our_approach} (see the lower left corner). In the Laplace domain, the expression for the admittance controller $Y(s)$ can be written as:
\begin{equation}
    Y(s) = \frac{V_{\mathrm{ref}}(s)}{F_{h}(s)} = \frac{1}{ms+b}
    \label{eq1}
\end{equation}

\noindent where $V_{\mathrm{{ref}}}(s)$ represents the corresponding reference velocity in the Laplace domain, which is provided by the admittance controller to be followed by the robot via its internal motion controller. $F_{h}(s)$ denotes the human force, while $m$ and $b$ represent the admittance mass and damping, respectively. The parameters of the admittance controller for each dof are the same and were carefully selected to maintain the stability of our pHRI system using the analysis approach suggested in \cite{aydin2021towards}.

Since the control loop ran at 500 Hz, our pHRI system could display real-time haptic feedback to users during the simulated tasks performed in the virtual environment as well as the tasks performed in the physical world. 

\subsection{Experiments}
    \label{sec:expA}

\subsubsection{Training Experiments}
\begin{figure}[t]
    \myVspace
	\includegraphics[width=\columnwidth]{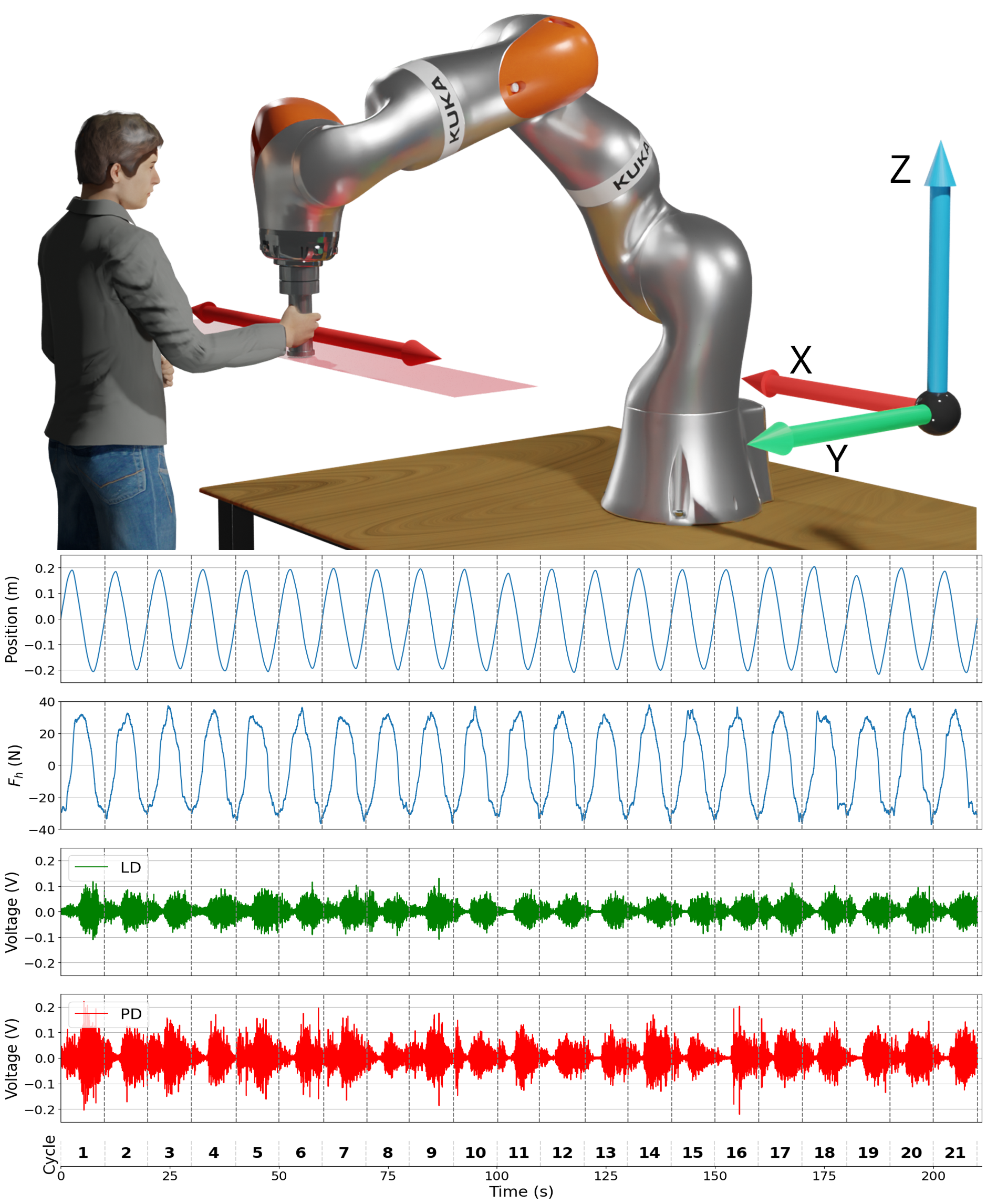}
	\centering
	\caption{The participants of our pHRI experiments were asked to perform cyclic and continuous movements along the X-axis. The end-effector position (first row), human force (second row), and surface EMG signals recorded from the LD (third row) and PD (fourth row) muscles until the onset of fatigue for a representative participant. The vertical dotted lines in the plots are used to separate the cycles.}
\label{fig:Data_From_One_Subject}
\end{figure}

The goal of the primary experiments is to collect data for training our learning models to estimate FCF. Participants were asked to hold the handle of the robot at the level of their umbilicus along the body's midline and repetitively perform translational left-right movements along the X axis (called lateral movements in the text) as shown in Fig.~\ref{fig:Data_From_One_Subject}. In each cycle of the movement, they moved the handle 20 cm to the left of the midline first, then 20 cm to the right of the midline, and finally back to the midline again. The admittance controller in the Y direction was disabled to constrain the robot’s motion to the XZ plane (frontal plane) only.

\begin{comment}
three sets of movements in our experiments as shown in \textcolor{red}{Figure XX} %\hyperref[fig:experiment-KUKA]{Figure \ref{fig:experiment-KUKA}}:
\begin{enumerate}[label=(\alph*)]
    \item linear cyclic movements in the transverse-plane along the X axis: they moved the handle to 15 cm of the left of midline and then 15 cm right to the midline and finally back to the midline (called left-right movements in the text) 
    \item linear cyclic movements in the sagittal-plane along the Y axis (called forward-backward in the text).
    \item circular cyclic movements on the transverse-plane (i.e. XY plane).
\end{enumerate}
\end{comment}

In each trial, participants were asked to follow a visual cursor making cyclic movements at 0.1 Hz on a computer monitor. The experiments were performed under two different admittance damping ($b$ = 275, 300 kg/s) to introduce variability to the data while the admittance mass was kept constant ($m$ = 10 kg).  

Before the experiments, participants underwent preparation for EMG electrode and sensor placement following the SENIAM recommendations \cite{hermens2000development}. The designated areas on their arms were shaved and cleansed with alcohol. A small amount of conductive gel was applied between the electrodes and the skin to reduce impedance and enhance signal acquisition. Based on the nature of our pHRI task detailed above, two EMG sensors were attached to the participants' arms to acquire data from the lateral deltoid (LD) and posterior deltoid (PD) muscles in our primary pHRI experiments.

In each trial, we measured the human force, the end-effector position of the robot, and surface EMG signals during the experiments. We also recorded the level of subjective rating of fatigue perceived (SRF) by the participant after each cycle and the number of cycles until the onset of fatigue (see Fig.~\ref{fig:Data_From_One_Subject}).

\begin{comment}
\begin{figure}[t]
    \myVspace
	%\includegraphics[width=\columnwidth]{figs/inference_workflow.jpg}
	\centering
	\caption{\textcolor{red}{Interaction force, robot end-effector position, and surface EMG signals from the MD and PD muscles recorded until the onset of fatigue for a representative participant.}}
	\label{fig:tasks}
\end{figure}
\end{comment}

\begin{comment}
\begin{figure}[t]
    \myVspace
	\includegraphics[width=\columnwidth]{figs/monitored_muscles.png}
	\centering
	\caption{Raw EMG signals acquired from Middle/Lateral Deltoid (MD), Posterior Deltoid (PD), Anterior Deltoid (AD), and Triceps Brachii (TB) muscles of the human arm during our pHRI task. The data is presented for 2 cycles of the movement only for clarity.}
	\label{fig:our_approach}
\end{figure}
\end{comment}

\begin{comment}
\begin{figure}[t]
    \myVspace
	\includegraphics[width=\columnwidth]{figs/monitored_muscles.png}
	\centering
	\caption{EMG electrodes and sensors were attached to 2 locations on the arms of participants to acquire data from Middle/Lateral Deltoid (MD) and Posterior Deltoid (PD). \textcolor{red}{SHOW THE LOCATION OF EMG SENSORS ON RENDERED VIEW OF HUMAN ARM. THE FIGURE HERE IS A SCREEN SHOT FROM DOGANAY'S STUDY, I PUT IT HER AS A REFERENCE NOTE THAT HE USED DIFFERENT MUSCLES}}
	\label{fig:our_approach}
\end{figure}
\end{comment}

\subsubsection{Participants}
Training experiments involved 10 participants with an average age of 22.5±1.9 years, an average height of 177±7.4 cm, an average body mass of 76.9±13.6 kg with a fat ratio of 15±3.6\%, and an average body mass index (BMI) of 24.3±2.2 kg/m². Each participant performed the task three times under each admittance damping (i.e. a total of 6 trials), with a five-minute rest period between repetitions. Hence, there were a total of 60 trials in the experiment (10 participants $\times$ 2 admittance dampings $\times$ 3 repetitions). The average number of cycles until the onset of fatigue was 23.3±6.9 and 24.2±7.6 under the admittance dampings of $b$ = 275 kg/s and 300 kg/s, respectively. 

Participants received both verbal and visual instructions on the placement of EMG electrodes and sensors, the robot, and the experimental protocol before the experiments. They completed two familiarization sessions on separate days before the actual experiment, during which they performed the task twice until the onset of fatigue. These training sessions were conducted to ensure that participants were adequately accustomed to the task before the actual trials.

\begin{comment}
\begin{table}[ht]
\centering
\caption{Experimental Conditions and Parameters}
\begin{tabular}{cccccc}
\toprule
\bottomrule
\end{tabular}
\label{tab:Experiment-Values}
\end{table}
\end{comment}

\subsection{Subjective Rating of Muscle Fatigue}
The Borg scale \cite{williams2017borg} was used to evaluate the rate of perceived exertion by the participants in our pHRI experiments. The Borg CR10 is a scale varying from 0 (no fatigue) to 10 (fatigue) and has been used in earlier studies in the domains of sports biomechanics, kinesiology, and rehabilitation. Participants received both verbal and visual instructions on the fatigue rating scale before the experiments. After each cycle of an experimental trial, they rated their fatigue level, starting from an initial score of ``0" and progressing to ``10," a degree at which point they could no longer continue the task.

\subsection{Machine Learning}
\subsubsection{Data Processing} 
The following features were extracted from the measured EMG signals to be used in the ML models: a) \emph{Frequency-Based Features:} MNF and MDF are frequently utilized in EMG studies. In addition, we used the total power of the signal (TP), which is the sum of the power spectral density over all frequencies, as the feature. b) \emph{Time-Based Features:} The RMS amplitude of the filtered and rectified EMG signal was used as a feature. 

Each cycle of an experimental trial was divided into 3 equal intervals, and the minimum and maximum values of the EMG features (MNF, MDF, TP, and RMS) were extracted for each muscle. For the extraction of frequency-based features, the raw EMG signals were filtered with a band-pass Butterworth filter having lower and upper cutoff frequencies of 5 Hz and 500 Hz, respectively \cite{sirintuna2020detecting}. Afterwards, the resulting signals were normalized by using Maximum Voluntary Contraction (MVC) muscle activation values. This enabled us to compare EMG activity in the same muscle on different days or in different individuals or between muscles. For extracting time-domain features, we then applied full-wave rectification by taking the absolute value of the normalized EMG signals. 
\begin{comment}
The dataset for each participant was divided into separate subsets for training and testing using the 80:20 split. Each model was developed using the training set, and its predictive performance was assessed on the test set. To further improve model generalization and mitigate overfitting, we employed a trial-wise cross-validation strategy, ensuring that data from each trial was systematically withheld for validation while the remaining trials were used for training.  
\end{comment}

\subsubsection{Modeling} 
We developed subject-specific ML models rather than a single model across all participants because inter-subject variability in EMG features and fatigue progression patterns is high, and personalized models captured each subject’s unique neuromuscular response to repeated exertion. In each model family, a separate model was built for each subject, trained on only that subject's data.

We trained and tested three regression models: Random Forest, XGBoost, and Linear Regression. Random Forest and XGBoost are based on decision trees and can capture non-linear relationships in regression. They are also efficient both in training and inference time compared to other model families. On the other hand, A linear regression model was utilized for the baseline comparison with Random Forest and XGBoost. 

The EMG features (MNF, MDF, TP, and RMS) extracted for each cycle were fed to our subject-specific ML models as input. %There were around 2208 EMG features used for training and testing the learning models of each participant: 6 trials $\times$ an average of 23 cycles per trial $\times$ 8 EMG features (including the minimum and maximum values per feature per cycle and channel) $\times$ 2 EMG channels. 
Features were extracted on a per-cycle basis, where each motion cycle—segmented according to position—was treated as a single data sample for the ML model. For each subject, 6 trials were performed with an average of 23 cycles per trial, yielding approximately 138 samples per subject. Each sample comprised a total of 16 features, derived from eight features (minimum and maximum values of MNF, MDF, TP, and RMS amplitude) computed for each of the two EMG channels across the cycle. The output was the fraction of cycles to fatigue (FCF). 
%Fatigue increases as the task progresses, so FCF and SRF are expected to be correlated, as observed in our study. 

A hyperparameter search was performed to identify the optimal configuration for each ML model. The Random Forest model was configured with 200 decision trees and a maximum depth of 8. The XGBoost model was configured with 200 boosting rounds and a maximum tree depth of 5, a learning rate of 0.05, and regularization parameters $\alpha = 0.5$ (L1) and $\lambda = 2.0$ (L2).

\subsection{Deep Learning}
We employed a CNN model for deep learning. Raw EMG signals were filtered first, subsequently normalized by the muscle activation values of MVC as described in the ML section, and then transformed into the frequency domain via Short-Time Fourier Transform (STFT) with 400-sample windows and 300-sample overlap, restricted to the 10–250 Hz frequency band. The resulting spectrograms were log-transformed and subsequently z-score normalized before being input to the model. The output was the fraction of cycles to fatigue (FCF).

A hyperparameter search was performed to identify the optimal configuration for the CNN model. The CNN model consisted of three convolutional blocks (32, 64, and 128 filters) with batch normalization and dropout (0.3), followed by fully connected layers of 128 and 64 units. Training employed the Adam optimizer (learning rate = 0.001), batch size = 16, and early stopping with a patience of 40 epochs. Data were partitioned into 80\% training and 20\% validation sets.

\subsection{Cross-Task Generalization}
To provide a proof-of-concept evaluation of the cross-task generalization capability of the proposed fatigue estimation models, additional experiments were conducted with a single participant: (i) a vertical movement task involving repetitive up-down movements along the Z-axis for a total distance of 40 cm, and (ii) a planar circular movement with a diameter of 20 cm on the XY plane (see the axes in Fig.~\ref{fig:Data_From_One_Subject}). For the vertical movement task, to test the robustness of the proposed approach not only to the movement direction but also the recruitment of different muscles, EMG signals were recorded from the anterior deltoid (AD) and lateral deltoid (LD) muscles, while the circular movement task utilized EMG signals from the same muscle pair (PD and LD) used in the primary dataset. The participant performed 6 trials again for each task. The data collected for additional tasks were not included in the training phase of the learning models and served as cross-task evaluations to assess the model’s robustness to changes in movement direction, arm kinematics, and muscle recruitment.

\section{Results}
    \label{sec:expAResults}

\subsection{EMG Features} 
\begin{figure}[t]
    \myVspace	\includegraphics[width=\columnwidth]{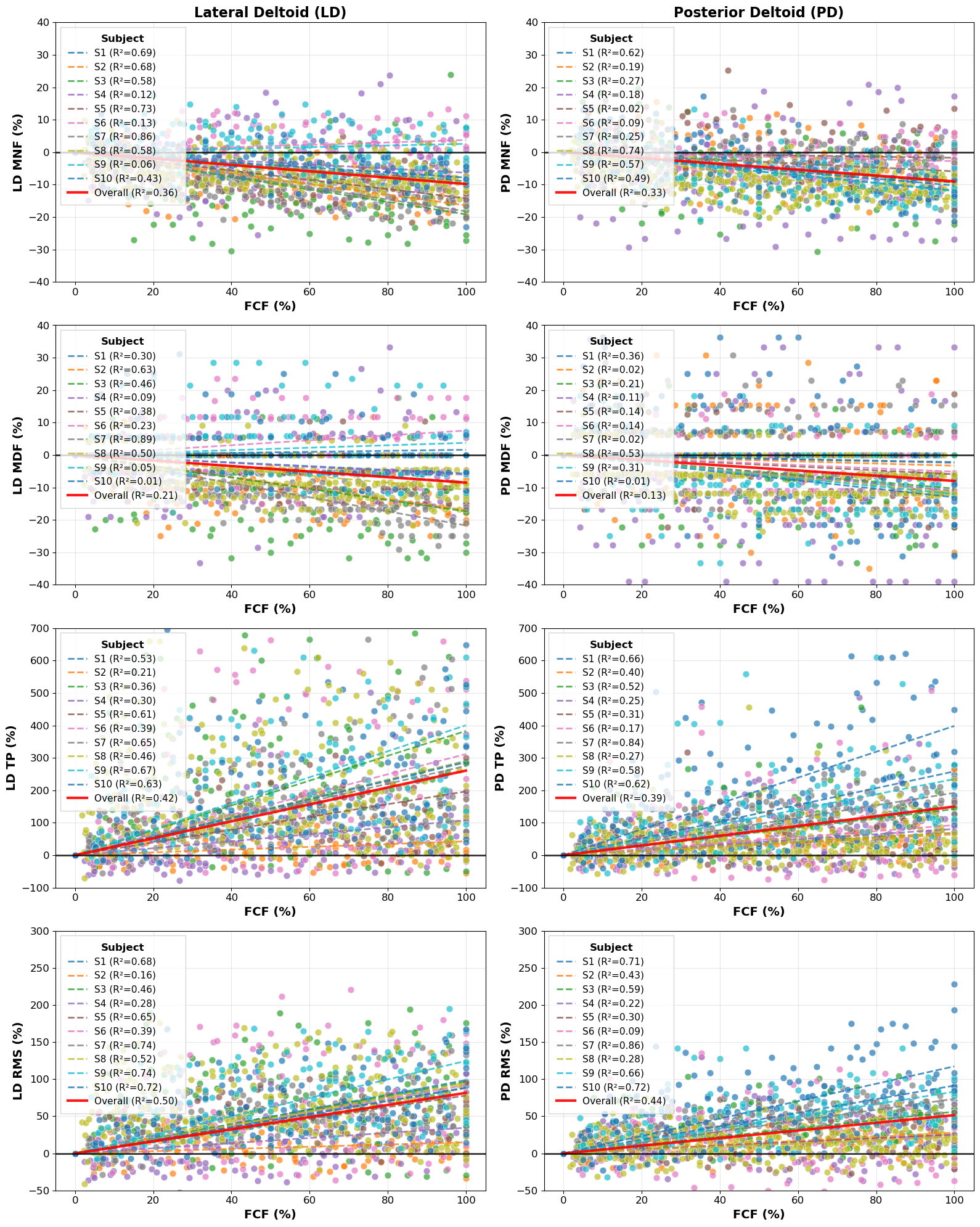}
	\centering
	\caption{Relative change in the maximum values of MNF, MDF, TP, and RMS amplitude with respect to the initial cycle plotted against the fraction of cycle to fatigue (FCF). Each row represents one feature, while the left and right columns are for the data collected from the LD and PD muscles. In each plot, a dot represents a feature for a cycle, and the data of each participant is represented by a unique color. The dashed lines in each plot represent the individual linear fits for each participant, while the solid line depicts the linear fit for the average across all participants to illustrate the general trend.}
	\label{fig:EMG_features}
\end{figure}

Fig.~\ref{fig:EMG_features} illustrates the percent change in EMG features relative to the initial cycle, plotted against the FCF. Given the variability in muscle activation levels among subjects and the differing number of cycles each subject completes before reaching fatigue, normalizing both the EMG data to the initial cycle and the number of cycles to the maximum cycles is essential. This approach enables the aggregation and comparison of data across participants within a unified framework. As shown in Fig.~\ref{fig:EMG_features}, MNF and MDF decrease with the number of cycles, while TP and RMS increase. The linear regression models were fitted to illustrate the overall trends between changes in EMG features and the fraction of cycles to fatigue (FCF). The considerable variation in R\textsuperscript{2} values across participants reflects inter-individual differences in EMG activation patterns, thereby supporting the need for subject-specific learning models.

\subsection{Subjective Rating of Participants} 
\begin{figure}[t]
    \myVspace	\includegraphics[width=\columnwidth]{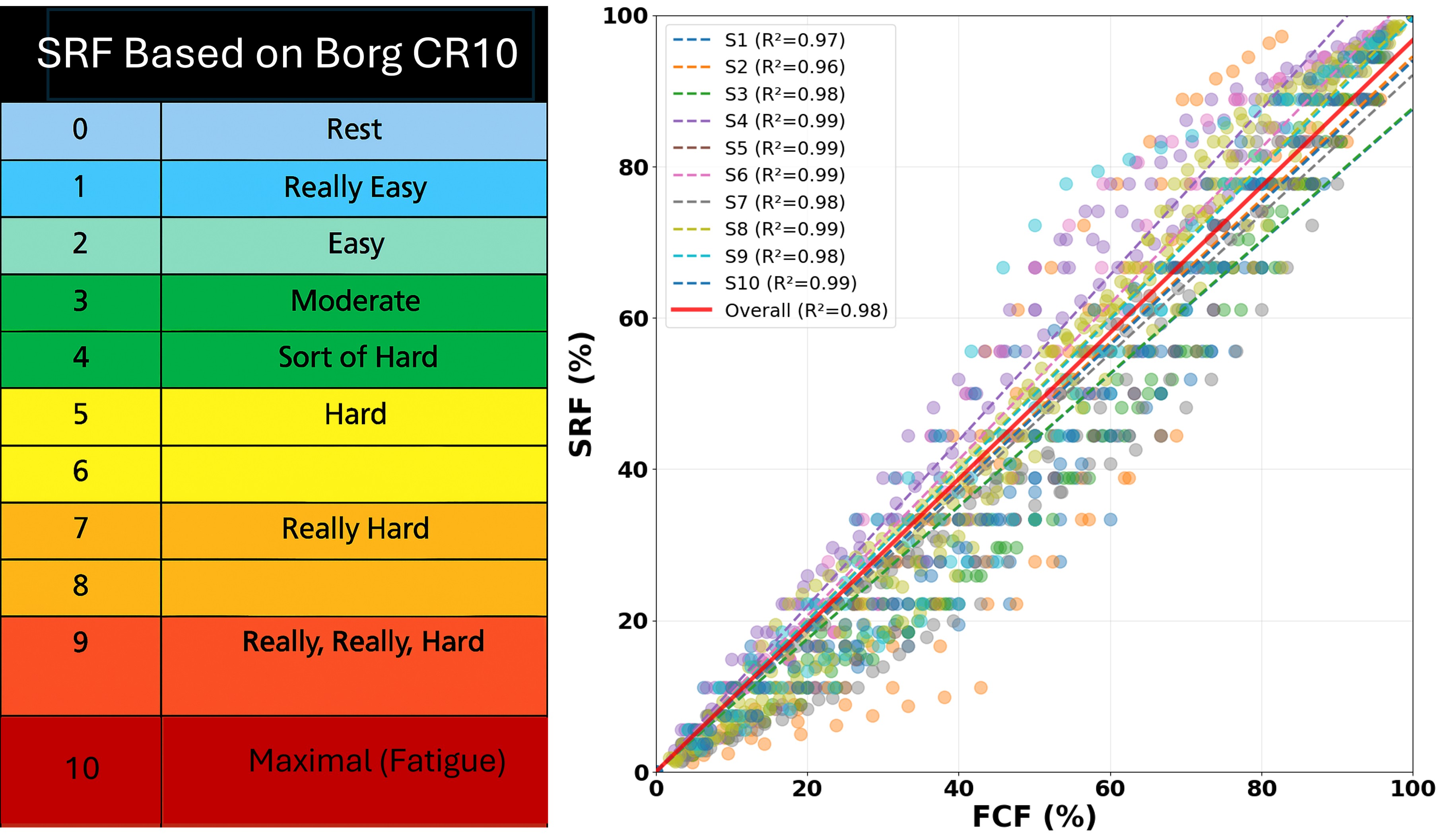}
	\centering
	\caption{a. Subjective level of fatigue perceived by the participants (SRF) was evaluated using Borg CR10 score. b. The fraction of cycles to fatigue (FCF) versus SRF. Each dot in the plot represents an SRF value of a participant recorded for a cycle. Data for each participant is represented by a unique color in the plot. The dashed lines represent the individual linear fits for each participant, while the solid line depicts the linear fit for the average across all participants to illustrate the general trend.}
	\label{fig:FCF_vs_SRF}
\end{figure}

Fig.~\ref{fig:FCF_vs_SRF} illustrates the subjective rating of fatigue (SRF) plotted against FCF. Although the original subjective scores were collected on a 0–10 scale in accordance with the Borg CR10 scale, the values were linearly interpolated and normalized to a percentage scale (0–100\%) to ensure compatibility with the fraction of cycles to fatigue (FCF) metric. As shown from the plot, there is a high correlation between FCF and SRF (R\textsuperscript{2} = 0.98), suggesting that the FCF value estimated by the learning models is a good indicator of fatigue perceived by the participant.

\subsection{Learning Models}

\begin{comment}
We investigated the effect of incorporating temporal history by including EMG features from the preceding N = 1, 3, and 5 cycles as additional input variables (note that N = 1 indicates that the EMG features for the current cycle are used to estimate the fatigue at the current cycle). This approach allowed us to evaluate whether incorporating recent historical context improves model accuracy by capturing the dynamic evolution of muscle fatigue over time.
\end{comment}
Root Mean Squared Error (RMSE) was employed as the primary metric to evaluate the performance of the ML and DL regression models in this study. Fig.~\ref{fig:Box_Plots_Participants} shows the mean RMSE values computed for each participant, averaged across all trials, for the respective regression models. Since the learning models output the estimated fatigue in percentage (FCF), RMSE values are also reported in percentage. Model performance was assessed using a trial-wise 6-fold cross-validation scheme, implemented as a Leave-One-Trial-Out (LOTO) procedure. Specifically, for each fold, the data from one of the six trials performed by a participant was held out for testing, while the remaining five trials were used for training. This procedure was repeated across all six possible trial selections. After each iteration, RMSE was calculated on the held-out trial, and the final performance was reported as the average RMSE across the six folds.
\begin{comment}
 In participant-wise cross-validation, the data from one participant (all 6 trials) was held out as the test set, while the model was trained on data from the remaining 7 participants in each fold. This process was repeated for all 8 participants.  Performance was assessed by computing the RMSE and R\textsuperscript{2} between the predicted and actual values on the held-out participant. The reported results represent the mean values of those metrics across the 8 folds. 
In trial-wise cross-validation, one trial was randomly selected from each of the 8 participants and held out for testing, resulting in 8 test trials per fold. The model was trained on the remaining 40 trials, and this process was repeated across XX folds with different random selections. After each iteration, RMSE and R\textsuperscript{2} were computed on the held-out trials. Final performance was reported as the mean values of those metrics averaged across the XX folds.   
\end{comment}

\begin{comment}
\begin{table}[ht]
\centering
\caption{Regression Performance of the ML Models Used in Our Study}
\begin{tabular}{cccccc}
\toprule
\bottomrule
\end{tabular}
\label{tab:regression-performance}
\end{table}
\end{comment}

As shown in Fig.~\ref{fig:Box_Plots_Participants}, the CNN model yielded the lowest RMSE values across participants, while the Linear Regression largest. When averaged over all participants, the RMSE was 20.8 ± 4.3\%, 23.3 ± 3.8\%, 24.8 ± 4.5\%, and 26.9 ± 6.1\% for the CNN, Random Forest, XGBoost, and Linear Regression models, respectively. 
Feature importance analysis using the Random Forest model indicated that MNF contributed the highest predictive relevance for estimating FCF (43\%), followed by RMS amplitude (22\%), TP (22\%), and MDF (14\%). A similar result was obtained by the feature importance analysis using the XGBoost model.

% \begin{figure}[t]
%     \myVspace	\includegraphics[width=\columnwidth]{figs/Box_Plots.png}
% 	\centering
% 	\caption{Box plots showing the change in R\textsuperscript{2} (a) and RMSE as a function of preceding cycles for the Random Forest Model trained and tested on the primary (seen) data (i.e. repetitive movement along the x-axis).}
% 	\label{fig:Box_Plots}
% \end{figure}

\subsection{Cross-Task Generalization}
\begin{figure}[t]
    \myVspace	\includegraphics[width=\columnwidth]{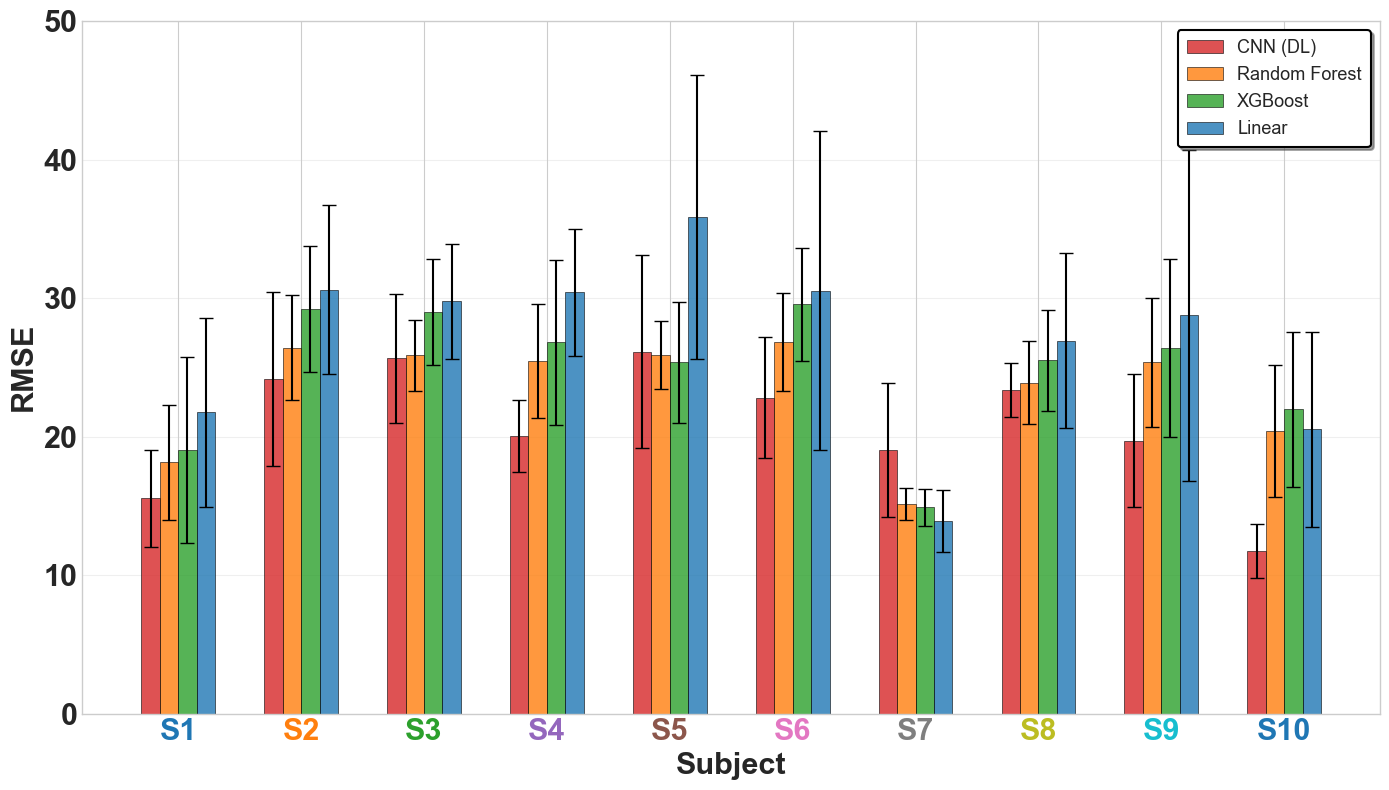}
	\centering
	\caption{Bar plots illustrating the mean RMSE values of each participant for the primary task, averaged across all trials, for the different regression models. }
	\label{fig:Box_Plots_Participants}
\end{figure}

\begin{figure}[t]
    \myVspace	\includegraphics[width=\columnwidth]{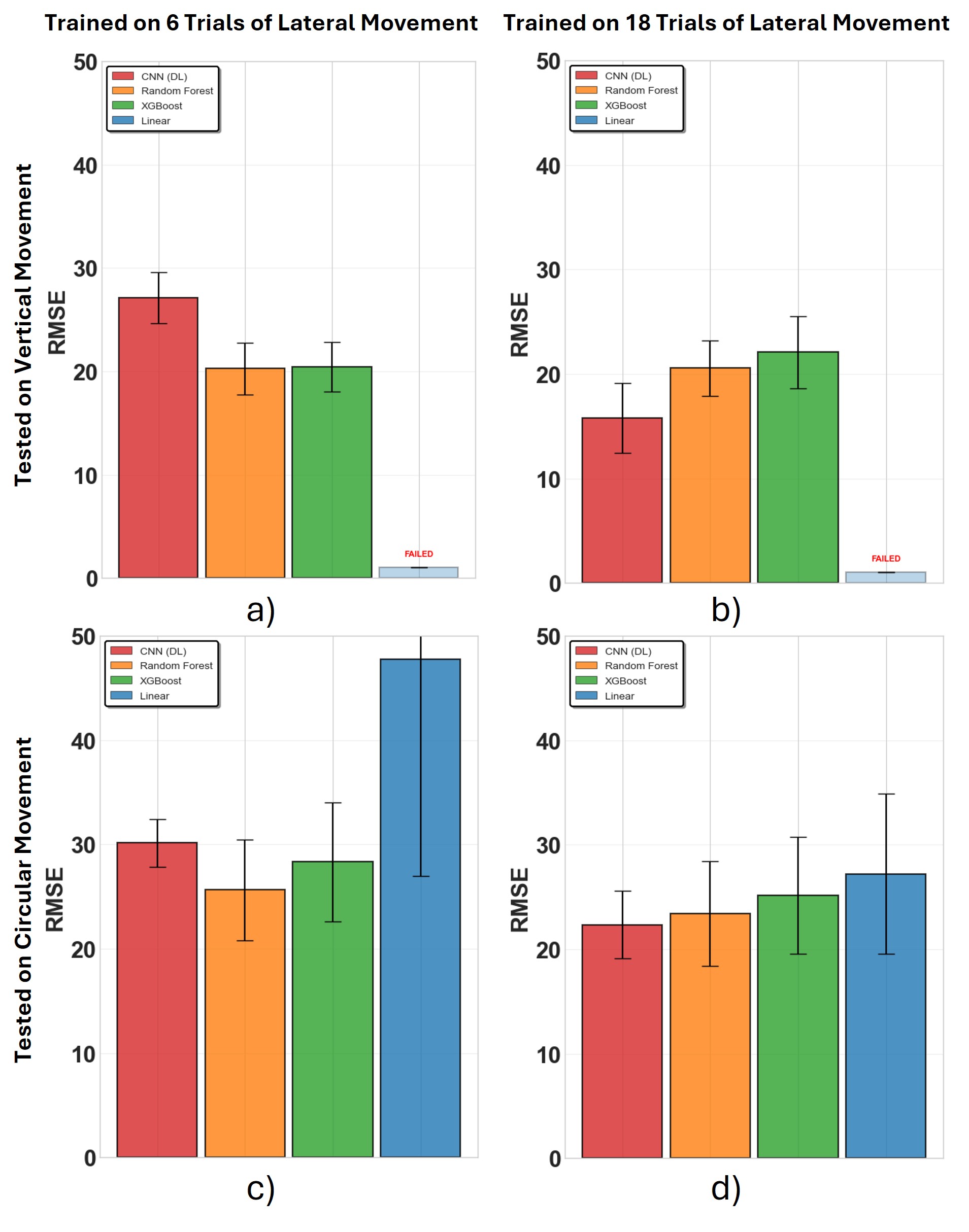}
	\centering
	\caption{Mean RMSE values for participant S1 in cross-task experiments. Panels (a,c) correspond to models trained on 6 lateral trials and tested on vertical and circular tasks, while panels (b,d) correspond to models trained on 18 lateral trials and tested on the same tasks.}
\label{fig:Box_Plots_Cross_Task}
\end{figure}

Cross-task generalization experiments were conducted with a single participant (S1 in Fig.~\ref{fig:Box_Plots_Participants}). Subject-specific learning models were trained on EMG data from the lateral movement task, initially using 6 trials and subsequently an extended dataset of 18 trials (6 original + 12 additional) to assess the effect of training set size on model performance. These models were then applied without retraining or fine-tuning to new EMG datasets obtained from two unseen tasks: (i) vertical repetitive movements along the Z-axis and (ii) circular movements on the XY-plane. Model generalization performance was quantified using the same regression metric (RMSE) employed in the primary experiments. The RMSE values for the vertical movement task trained on 6 trials (Fig.~\ref{fig:Box_Plots_Cross_Task}a) were 27.1 ± 2.5\% for CNN, 20.3 ± 2.5\% for Random Forest, and 20.4 ± 2.4\% for XGBoost. When the training set was extended to 18 trials (Fig.~\ref{fig:Box_Plots_Cross_Task}b), the corresponding RMSE values changed to 15.8 ± 3.4\%, 20.6 ± 2.7\%, and 22.1 ± 3.5\%, respectively.
For the circular movement task, the RMSE values obtained with 6 training trials (Fig.~\ref{fig:Box_Plots_Cross_Task}c) were 30.1 ± 2.3\% for CNN, 25.6 ± 4.9\% for Random Forest, 28.3 ± 5.7\% for XGBoost, and 47.8 ± 20.8\% for Linear Regression. With the training set expanded to 18 trials (Fig.~\ref{fig:Box_Plots_Cross_Task}d), the RMSE values changed to 22.4 ± 3.2\%, 23.4 ± 5.0\%, 25.1 ± 5.6\%, and 27.2 ± 7.7\%, respectively. Note that the Linear regression model trained with the data of either 6 or 18 trials of the lateral movement failed to regress (i.e. generated an average RMSE value larger than 100\%) when tested on the data of the vertical movement task. It also generated high RMSE values when tested on the circular movement task.

\section{Discussion and Conclusion}
    \label{sec:conclusion}
This investigative study introduces a learning-based approach to estimate the progression of human effort towards muscular fatigue in dynamic and repetitive physical human-robot interaction (pHRI) tasks, based on surface electromyography (sEMG) signals. Earlier data-driven ML/DL methods for fatigue estimation largely relied on classification-based formulations. In contrast, our regression-based approach models the continuous progression of fatigue, allowing for closer monitoring of physiological changes and supporting timely intervention and adaptive control in dynamic tasks.

The results of our study show that both frequency-domain (MNF, MDF, TP) and time-domain (RMS amplitude) features of EMG signals evolve systematically with the progression of task (Fig.~\ref{fig:EMG_features}). The trends observed in EMG feature progression (decreasing MNF and MDF and increasing RMS and TP) reflect underlying physiological mechanisms such as decreased muscle fiber conduction velocity and increased motor unit recruitment as suggested in \cite{merletti1990myoelectric}. On the other hand, no significant changes in human-applied force were observed across cycles during task progression (see the second row of time-plots in Fig.~\ref{fig:Data_From_One_Subject}), indicating that participants consistently generated sufficient force to track the desired trajectory despite changes in their muscle activity. Among the EMG features, MNF appeared to have the highest predictive power for estimating the fraction of cycles to fatigue (FCF). The strong correlation observed in our study (Fig.~\ref{fig:FCF_vs_SRF}) between subjective fatigue rating (SRF) and FCF validates the use of FCF as a reliable metric for estimating fatigue in cyclic pHRI tasks. 

Our study demonstrates that constructing a cross-subject ML/DL model for fatigue estimation is challenging due to substantial inter-subject variability arising from physical and physiological differences. This variability was evident in the spread of R\textsuperscript{2} values obtained when fitting EMG features to the fatigue cycle fraction (FCF) (Fig.~\ref{fig:EMG_features}). Notably, participants exhibited different activation strategies for the LD and PD muscles, with some relying more heavily on LD while others engaged PD to a greater extent. Consequently, we adopted subject-specific learning models for fatigue estimation. Even at the subject level, the root mean square error (RMSE) of FCF estimation ranged from 12\% to 26\%, averaging 21\% across participants for the best model (CNN). This error magnitude is attributable primarily to the dynamic nature of the task, which complicates acquisition of high-quality EMG signals compared to isometric contractions under static loading. Additional intra-subject variability in EMG features was observed across trials, although this was less pronounced than inter-subject variability. Performance across different ML models was comparable: Random Forest and XGBoost both achieved an RMSE of approximately 24\%, while even the Linear Regression model yielded similar accuracy of 27\%, though the standard deviation from the mean was the largest. 
%This outcome may suggest that the temporal evolution of normalized EMG features over successive cycles exhibits limited nonlinearity.

Prior studies have predominantly focused on static tasks involving isometric contractions or used computationally intensive musculoskeletal simulations, limiting real-time applicability. This work advances the state-of-the-art by proposing a data-driven solution for dynamic tasks using lightweight learning models that can be deployed for real-time fatigue estimation. Unlike previous efforts, which required musculoskeletal models or ODEs for modeling the force-fatigue relation, our method offers a simpler implementation. By inputting the percent relative change in EMG features with respect to the first cycle rather than their absolute counterparts into the ML models, we account for intra-subject variability in baseline muscle activation and endurance capacity, thereby enhancing the estimation capacity of the subject-specific learning models, especially in cross-tasks. 

In cross-task validation experiments (Fig.~\ref{fig:Box_Plots_Cross_Task}), we demonstrated that the ML models (Random Forest and XGBoost) trained with the relative changes in EMG features are robust to changes in motion kinematics (lateral vs vertical and circular movements) and muscle recruitment (data was collected from LD and PD muscles in lateral and circular movement tasks, while LD and AD for the vertical movement task). When the ML models were trained on either 6 or 18 trials of the lateral movement data, their test performance on the vertical and circular tasks showed little degradation, likely because both training and testing phases used EMG features normalized relative to the first cycle. In contrast, the CNN model yielded a higher average RMSE than the ML models when trained on 6 trials (Figs.~\ref{fig:Box_Plots_Cross_Task}a and \ref{fig:Box_Plots_Cross_Task}c), possibly due to differences in the frequency content of the training and testing EMG signals. However, its performance improved when trained on 18 trials of the lateral movement (Figs.~\ref{fig:Box_Plots_Cross_Task}b and \ref{fig:Box_Plots_Cross_Task}d), as the larger dataset provided more representative EMG patterns and enhanced generalization across tasks.

Some limitations of the proposed approach should be acknowledged here. First, the dataset was collected from a specific group of participants who were university students, which may limit generalizability across populations with different physical and physiological characteristics. Second, the model assumes homogeneous fatigue progression across different muscles, although muscle volume, type, and activation patterns may differ.  

Future research should explore integrating EMG signals with joint kinematics and external force measurements to develop hybrid models that leverage the physical interpretability of musculoskeletal simulations (e.g., using OpenSim) with the predictive power of data-driven methods. Including diverse tasks, body segments, and real-world scenarios (e.g., manufacturing, rehabilitation) would improve robustness. Additionally, closed-loop robotic controllers that adapt based on real-time fatigue estimates can be developed and validated in longitudinal studies.

\begin{comment}
{Acknowledgment}
P.P.N. and A.M. acknowledge the research fellowship provided by the KUIS AI Center. We also acknowledge the technical support provided by Dr. Yusuf Aydin in the early stages of this work. The raw EMG signals, extracted EMG features, and corresponding subjective ratings of fatigue (SRFs) from the primary experiment involving ten participants are publicly available at:
\href{https://github.com/Kiki-robots/Fatigue_Estimation}{EMG Fatigue Estimation Dataset}.
\end{comment}

\ifCLASSOPTIONcaptionsoff
  \newpage
\fi

\bibliography{References_fixed}

\end{document}